\documentclass[11pt]{article}
\usepackage{coling2020}
\usepackage{times}
\usepackage{url}
\usepackage{latexsym}
\usepackage{graphicx}
\usepackage{color}
\usepackage{float}
\colingfinalcopy 

\title{Fact-level Extractive Summarization with \\ Hierarchical Graph Mask on BERT}

\author{Ruifeng Yuan \quad Zili Wang \quad Wenjie Li\\
  Department of Computing, The Hong Kong Polytechnic University, Hong Kong\\
  Xidian University \\
  {\tt \{csryuan, cswjli\}@comp.polyu.edu.hk} \\
   \tt  ziliwang.do@gmail.com \\}
 
\date{}

\begin{document}
\maketitle
\begin{abstract}
	Most current extractive summarization models generate summaries by selecting salient sentences. However, one of the problems with sentence-level extractive summarization is that there exists a gap between the human-written gold summary and the oracle sentence labels. In this paper, we propose to extract fact-level semantic units for better extractive summarization. We also introduce a hierarchical structure, which incorporates the multi-level of granularities of the textual information into the model. In addition, we incorporate our model with BERT using a hierarchical graph mask. This allows us to combine BERT’s ability in natural language understanding and the structural information without increasing the scale of the model. Experiments on the CNN/DaliyMail dataset show that our model achieves state-of-the-art results. \blfootnote{The code is available at \url{https://github.com/Ruifeng-paper/FactExsum-coling2020}}

	\blfootnote{This work is licensed under a Creative Commons Attribution 4.0 International License. License details: \url{http://creativecommons.org/licenses/by/4.0/.}}
\end{abstract}

\section{Introduction}

The aim of text summarization is to generate a compressed shorter highlight of a given document. The generated summaries must conform to natural language constraints and cover the most important information conveyed in the source text. In this paper, we focus on extractive summarization which generates summaries by selecting salient sentences or other semantic units from the source text.

As we know, sentences are regarded as the basic textual units making up the final summaries in most extractive summarization models. One of the problems with sentence-level extractive summarization is that there exists a gap between the human-written gold summary and the training objective. Since most datasets do not contain sentence-level labels, oracle summaries are normally generated by a greedy algorithm, which maximizes the ROUGE score between the sentences in the source text and in the gold summary \cite{nallapati2017summarunner,liu2019text}. As pointed out by \cite{narayan2018ranking}, sentences with high ROUGE scores may not necessarily lead to an optimal summary. Such discrepancy may be due to the overlapping contents and over extraction. Therefore, some researches try to perform extraction at a lower level such as words or phrases \cite{bui2016extractive,cheng2016neural,gehrmann2018bottom}. Although these models are able to learn which phrases or words are more important directly from the gold summary, it is hard to maintain semantic integrity when information is scattered.

Inspired by \cite{cao2018faithful}, we propose to extract facts, a granularity between phrases and sentences, as the primary textual units to generate summaries. In this work, we first develop a heuristic algorithm to split a sentence into several smaller units, where each unit is considered as a single fact description. Then we apply a one-to-one strategy to match each fact in the gold summary to one fact in the source text to obtain the oracle label. In this way, we smooth the gap between the gold summary and oracle labels and further reduce redundancy and over-extraction. Meanwhile, fact units can still well keep semantic integrity and thus are able to provide faithful summaries. Considering facts are relatively independent semantic units, in order to maintain rich contextual information and to capture potential relationships among facts, we believe the sentence-level information still plays an important supporting role. Moreover, based on the assumption that the selected facts should be semantically close to the majority of source documents, we also import the document-level information as part of our model. We will show that such a fact-sentence-document hierarchical text modeling facilitates to capture more structured contextual information for effective fact extraction.

Recent works \cite{liu2019text,bae2019summary,zhang2019hibert} have demonstrated that it is highly beneficial to apply pretrained language models such as BERT to extractive summarization models. Following this trend, we adopt a BERT-based encoder to generate contextualized representations for further extraction. It is challenging to impose the structure information when fine-tuning the BERT model in a downstream task.  \cite{fang2019hierarchical,zhang2019hibert} separately encode representations for each granularity with BERT and then capture the structure information with a graph network stacked upon the encoder. However, this method not only results in a large complex model but also cannot fully utilize the advantage of pretrained language models. Based on the idea that BERT can be regarded as a full connected graph, we propose to directly impose the structure information on BERT with a graph-based mask to jointly learn contextual representations of different text granularities within a single BERT. On the one hand, we significantly reduce the time cost of running BERT several times plus running a graph neural network to running BERT once. On the other hand, we are able to utilize the complex structure of BERT to capture the structure information.

We conduct experiments on the CNN/DailyMail dataset, a well-known extractive summarization benchmark dataset. The results show that using facts as extractive units improves the summarization quality and our model achieves better performance than the state-of-the-art model. 
The contributions of our work can be summarized as follows.
(1) We propose to extract fact-level semantic units for extractive summarization to fill the gap between the gold summary and oracle labels, reducing redundancy and over-extraction. (2) We import a hierarchical structure to remedy the information loss after splitting sentences into facts. (3) We propose a graph-based mask algorithm to impose the structure information on BERT directly. To the best of our knowledge, we are the first to combine the pretrained language model and the structure information without increasing the scale of the model, which provides a new idea for fine-tuning pretrained models on downstream tasks. 

\section{Related Work}

\subsection{Extractive Summarization}
Extractive summarization attempts to select the most important sentences from the source text and subsequently concatenate them as a summary. With neural network models, researchers usually formulate it as a sentence binary classification problem. SummaRuNNer \cite{nallapati2017summarunner}, one of the earliest neural summarization models, applies an RNN-based encoder to generate sentence representations and a neural classifier to determine which sentences should be included in the summary. \cite{narayan2018ranking} further extends SummaRuNNer with a reinforcement model, which optimizes the summary-level ROUGE metric. Some other works achieve better performance through more complex models. \cite{zhou2018neural} proposes a decoder that jointly learns to score and select sentences, while \cite{zhang2018neural} presents a latent variable extractive summarization model, which directly maximizes the likelihood of human summaries. In recent works, models based on pretrained models \cite{liu2019text,bae2019summary,zhang2019hibert}, especially BERT, have made a step forward.

\subsection{Graph Based Summarization}
Graph-based summarization methods aim to utilize the structure information of the text. Based on the assumption that important sentences are often linked with each other, unsupervised models like TextRank \cite{mihalcea2004textrank} and LexRank \cite{erkan2004lexrank} show great ability to identify the salient information. \cite{liu2018toward} proposes an abstractive summarization framework based on the Abstract Meaning Representation (AMR) graph, which captures the propositional semantic information. \cite{koncel2019text} presents a graph transformer to generate one-sentence summaries from a knowledge graph. Meanwhile, other researches focus on learning latent tree structures while optimizing summarization models. \cite{williams2018latent} regards the tree structure learning problem as a reinforcement learning problem and \cite{liu2019single} generates a multi-root dependency tree where each root is a summary sentence.

\subsection{Pretrained Language Models}
Pretrained language models \cite{radford2018improving,devlin2018bert} have been proved to be extremely successful for making great progress in various nature language tasks. These models are originated from the idea of word embeddings \cite{pennington2014glove} and further extend it by pretraining a sentence encoder on the huge unlabeled corpus using language modeling objectives. Bidirectional Encoder Representations from Transformers (BERT) \cite{devlin2018bert}, one of the state-of-art language models, is trained with a masked language model and a next-sentence-predicting task. Recently, pretrained language models have been widely used to improve performance in language understanding tasks \cite{dong2019unified}. As for the extractive summarization task, it provides the powerful sentence embeddings and the contextualized information among sentences \cite{zhong2019searching}, which have been proven to be critical to extractive summarization.

\section{Extraction of Facts and Alignment with Gold Summary}

We propose to explicitly align a gold summary with its corresponding facts descriptions in the source text. For this purpose, we develop a heuristic algorithm to break up the source text and the summary into smaller granularities as introduced below.

When splitting sentences into smaller semantic units, we need to ensure each unit has a proper size while maintaining the integrity of information. An intuitive idea might be to split sentences with commas and other conjunctions. But this straightforward strategy is not capable of handling complicated sentences in documents. Therefore, we leverage the dependency parser to handle this issue. To begin with, we adopt a dependency parser to convert a sentence into the labeled tuples in the form of (word1, word2, label), where the label denotes a grammatical relation between a pair of words. After that, the sentence is split using the labels representing punctuation marks, conjunctions and clauses, including punct (punctuation marks), cc (coordination relationship) and mark (finite clause). To acquire a more complete semantic unit, we merge units that are connected by some special labels such as the relative clause modifier (acl:relcl), the adverbial clause modifier (advcl), the appositional modifier (appos) and the clausal complement (ccomp). Furthermore, we use the conjunct (conj) to identify whether a conjunction connects two sentences or two phrases. A conjunct is a relation between two elements connected by a coordinating conjunction. When the distance between the two elements is less than a threshold, we regard the two coordinated elements are words or phrases rather than clauses. If so, the two split parts are merged back as one unit.

Moreover, we predefine a minimum unit length and a maximum clause length. If the length of a unit is smaller than the minimum unit length, this unit will be merged with the preceding unit. If a clause is longer than the maximum clause length, we regard it as an independent semantic unit. Take the dependency trees in Figure 1 as an example. The sentence is split into five parts based on the above-mentioned rules. We merge the first three parts back together since they are connected by the appos label while not exceeding the threshold length for being an independent clause. As for the last two parts, although there is a coordination relationship, it turns out that they are two coordinate entities based on the conj label. Hence, we also merge the last two parts as a fact.

\begin{figure*}[t]
	
	\centering
	\includegraphics[scale=0.41]{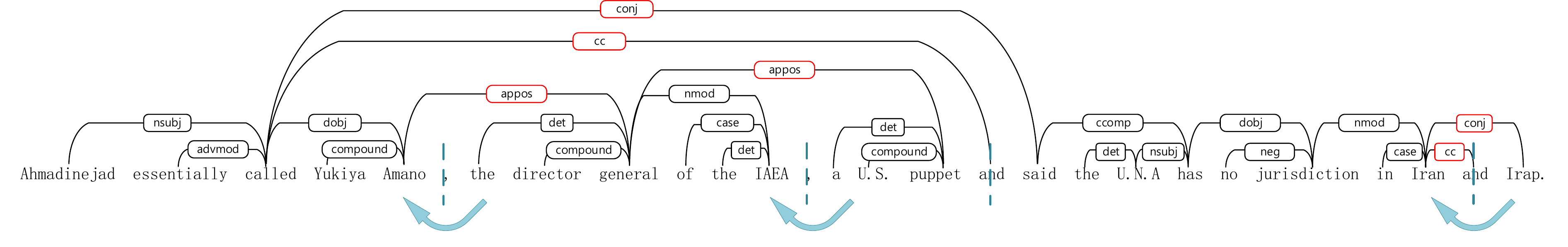}
	\caption{A dependency tree example. We extract the following two fact descriptions: Ahmadinejad essentially called Yukiya Amano, the director general of the IAEA, a U.S. puppet $|||$ said the U.N.A has no jurisdiction in Iran and Irap}
	\label{fig:label1}
	
\end{figure*}

In the experiments, we adopt the well-known NLP pipeline Stanford CoreNLP \cite{hermann2015teaching} for dependency parsing. Table 1 presents the statistics on the train set of CNN/DaliyMail. On average, a sentence contains 1.6 facts. It should be noted that some sentences remain as single facts, while others are split into 2 or 3 facts.

\begin{table}[H]
	\centering
	\begin{tabular}{lcc}
		\hline \textbf{granularity} & \textbf{num} & \textbf{len} \\ \hline
		Sentence & 34.3 & 24.7  \\
		Fact      & 51.5 & 14.8  \\
		\hline
	\end{tabular}
	\caption{\label{font-table}The average unit number and unit length in train set of CNN/DM dataset . }
\end{table}

Once facts are extracted, the next step is to obtain oracle labels for them from the gold summary. We match each fact in the summary with one corresponding fact that has the maximum ROUGE score from original text. Such an alignment allows each part of the summary to be accurately mapped to a semantic unit in the source text.

\section{Summarization Model}
\subsection{Model Framework}
As illustrated in Figure 2, our model consists of three components: a BERT-based encoder, a hierarchical graph mask, which incorporates the structure constraint on BERT and a classifier taking the multi-granularity information as input to extract salient facts to form the summary.

\subsection{BERT-based Encoder}
We use a BERT-based encoder to generate contextualized representations of the semantic units with different granularities. Since the outputs of BERT are grounded to tokens, we adapt the strategy that is similar to \cite{liu2019text} to modify the embeddings and the input sequence of BERT.

To obtain representations of the facts, the sentences and the document, we add a special token at the begining of each semantic unit. As shown in Figure 2, [cls\_d], [cls\_s] and [cls\_f] are inserted at the beginning of the document, the head of each sentence and the fact representing the embeddings of different granularities, respectively. Initially, these three kinds of [cls] tokens share the same pre-trained [cls] embedding. We also add a [seq] token at the end of each fact to separate the smallest semantic units. In addition, segment embeddings are used to identify multiple granularity levels within a document. For the $i$th granularity level, we assign the segment embedding $E_A$ or $E_B$ conditioned on whether $i$ is odd or even. For example, in Figure 2, we first assign the [cls\_d] segment embedding $E_A$ and then for the next granularity level, sentence, we allocate all [cls\_s] with $E_B$. With two different segment embeddings that 
separate adjacent granularity levels, the model is aware of the hierarchical structure among different granularity levels. 

\begin{figure}[t]
	
	\centering
	\includegraphics[scale=0.75]{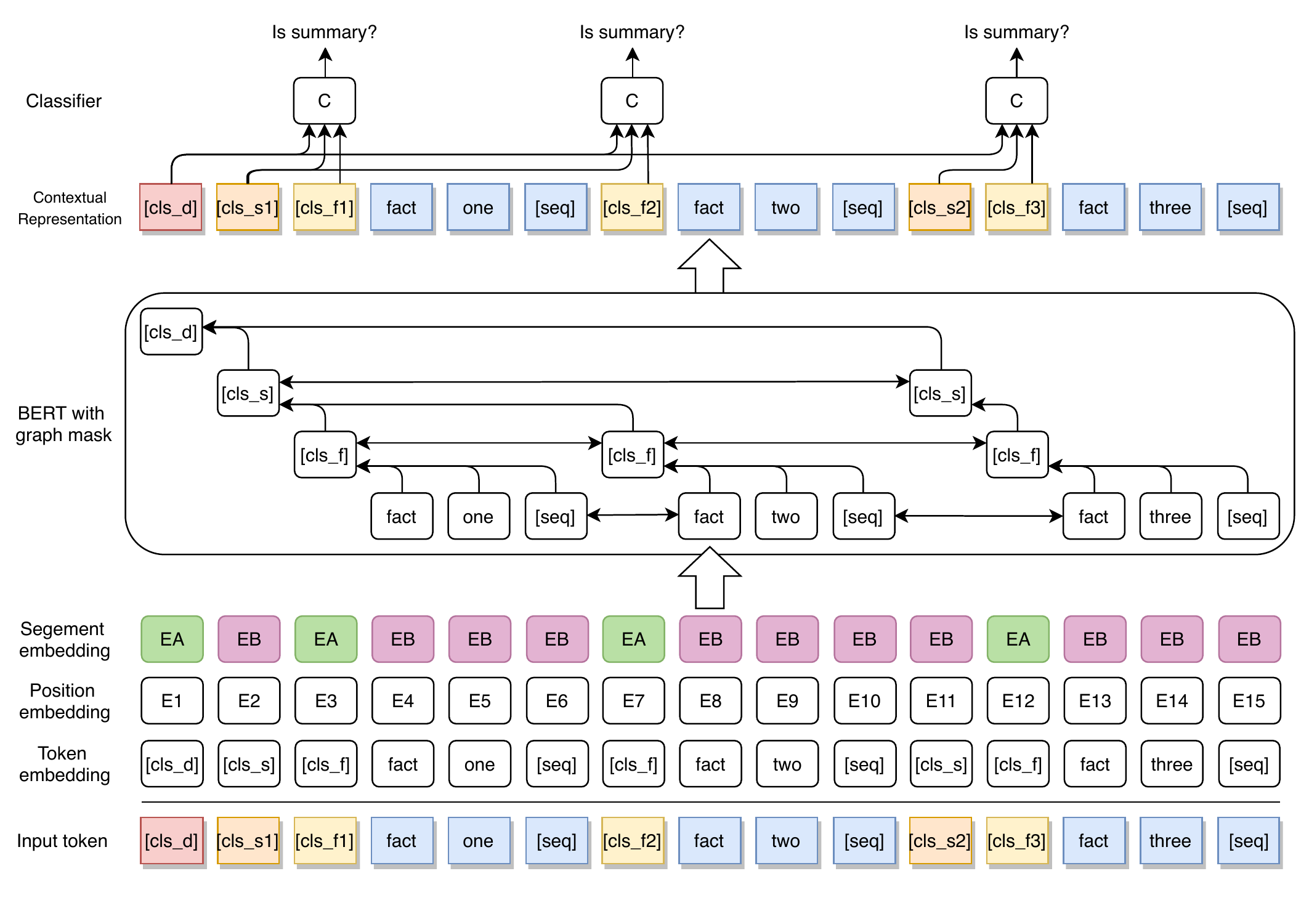}
	\caption{The framework of the summarization model}
	\label{fig:label1}
	
\end{figure}
\subsection{Hierarchical Graph Mask on BERT}
A document is composed of a collection of sentences, while each sentence may contain multiple facts. Such a hierarchical structure motivates us to generate representations for different granularity levels with hierarchical constrains on BERT. We propose a hierarchical graph mask to restrict information dissemination in BERT. To build a hierarchical graph, we add three kinds of edges to a token in the input sequence, including the bi-directional edges connecting to all other tokens at the same granularity level, the edges from the tokens belonging to them at the smaller granularity level and the edges to their corresponding tokens at the larger granularity level. For example, as shown in Figure 2, a fact token [cls\_f1] can disseminate its information to other fact tokens and the sentence token [cls\_s1], while receiving the information from the word tokens “fact”, “one” and [seq]. The tokens at each granularity level thus capture semantics from different sources. As a result, the hierarchical graph can effectively utilize all levels of structural information with different granularities. 

To implement this graph structure on BERT, we assign a mask vector to each token based on the hierarchical graph $H$. Given an input sequence with $n$ tokens $[t_1,t_2,...,t_n]$, the mask vector of $t_i$ is denoted by $[h_{i1},h_{i2},...,h_{in}]$, where $h_{ij}$ refers to whether there is a directed edge from the token $j$ to the token $i$ in the graph $H$. Taking [cls\_f1] in Figure 2 as an example, the mask vector of this token is $[0, 0, 1, 1, 1, 1, 1, 0, 0, 0, 0, 1, 0, 0, 0]$. After obtaining the mask vector for each token, we stack them to form a $n \times n$ mask matrix $M$ and calculate attentions with the equation below. For simplification, we write it in the one-head form.

$$
Attention(Q, K, V)=Softmax(\frac{QK^{T}M}{\sqrt{d^k}})V
$$

\noindent where $Q$ refers to the query matrix, $K$ refers to the key matrix, $V$ refers to the value matrix and $\sqrt{d^k}$ represents a scaling factor.

\subsection{Classifier}
After applying the hierarchical graph mask on BERT from words to document, we obtain representations of all levels of granularities. Although our focus is to extract the salient facts from a given document, we believe that the sentence-level information also contributes to identifying the importance among facts. Considering facts are rather independent semantic units, the sentence information is able to build up connections from a higher-level perspective and help locate the important facts more accurately. Moreover, we postulate that the selected facts should be semantically representative of the source document. Therefore, we also import the document representation as part of input to the classifier.

Given a fact representation $f_i$ from the last layer of BERT, we concatenate it with the document representation $d$ and its corresponding sentence representation $s_j$. Then we employ a sigmoid classifier to predict whether or not it should be selected into the summary:

$$
y^{'}_i=\sigma (W_{c}(d||s_j||f_i)+b_{c})
$$

\noindent where $||$ represents concatenation, $W_{c}$ and $b_{c}$ are the parameters of the classifier.

\subsection{Learning}

We apply a binary classification entropy of prediction $y^{'}_i$ against gold label $y_i$ as the training loss. Similar to \cite{liu2019text,vaswani2017attention}, we use the Adam optimizer with $lr = 2e^{-5}$, $\beta_1 = 0.9$ and $\beta_2 = 0.999$, and adopt a learning rate schedule with warming-up=10000:

$$
lr=2e^{-3}\cdot min(step^{-0.5},step\cdot warmup^{-1.5})
$$

For testing, we first predict the score $y_i$ for each fact using our model and then rank these facts with their scores to select the top-$4$ facts as the summary, considering the average number of facts contained in gold summaries is four. We also apply the Trigram Blocking algorithm \cite{paulus2017deep} to reduce redundancy. In particular, given the current summary $s$, a candidate fact $c$ will not be included in $s$ if there is a trigram overlapping between $c$ and $s$. In this way, we minimize the similarity between the current summary and the candidate sentence to increase information diversity and richness.

\section{Experiments}

\subsection{Dataset and Evaluation Metric}
We conduct experiments on the non-anonymized version of the CNN/DailyMail dataset \cite{hermann2015teaching}. The dataset is composed of news articles and their corresponding highlights that give brief overviews of articles. We apply the standard splits for training, validation, and testing (CNN: 90,266/1,220/1,093 and DailyMail: 196,961/12,148/10,397). Due to the length limitation of BERT, we follow the common practice to truncate all input documents to 512 tokens.

To evaluate the summarization quality, we apply ROUGE \cite{lin-2004-rouge} for automatic evaluation. Following the convention, we report ROUGE-1 (unigram), ROUGE-2 (bi-gram) and ROUGE-L (longest common subsequence) F1-score in the following experiments. Moreover, we manually evaluate the fact-level recall, precision and F1-score between system-generally summaries and human-written gold summaries.

\subsection{Details}
We build our model using PyTorch and the BERT-base-uncased version of BERT. All the input documents are tokenized by BERT's sub-words tokenizer. We train the model for at most 50000 steps and the batch size in each step is 32. After training for 10000 steps, the model is saved and evaluated for every 1500 steps. With the three best checkpoints on the validation dataset, we record best model on the test dataset among the three.

\subsection{Automatic Evaluation}
We first examine the ROUGE results of the oracle summaries matched with different granularities. As shown in Table 2, by aligning facts with gold summaries it achieves significant improvement on all three ROUGE metrics. It shows that our idea of building oracle summaries based on facts is able to generate more accurate oracle labels for training. 

\begin{table}[H]
	\centering
	\begin{tabular}{lccc}
		\hline \textbf{Granularity} & \textbf{R-1} & \textbf{R-2} & \textbf{R-L} \\ \hline
		Fact & 57.96  & 34.93 & 54.69\\
		Sentence & 52.59 & 31.24 & 48.87 \\ \hline
	\end{tabular}
	\caption{\label{font-table}Results on oracle summary on CNN/DM test set.}
\end{table}

Table 3 summarizes the results of a variety of models on the CNN/DailyMail test set. The first section in the table presents two lead-3 baselines that select the first three sentences as summaries. One is our own implantation, while the other one is copied from \cite{liu2019text}. The slight discrepancy on ROUGE scores may be because of the difference in data preprocessing.

The second section in the table displays performances of the existing extractive models, including the baseline model SummaRunner, two SOTA models with no pretraining and several models based on BERT. All the results are directly taken from their respective papers. 

The last section reports the results of our own implementations. In order to avoid interference of other factors, we implement the SOTA model BERTSUM in our training environment. Meanwhile, for better comparison, we propose a fact-extraction BERTSUM baseline, which simply changes the primary textual unit for extraction from sentences to facts under the same architecture as BERTSUM. Compared to BERTSUM, using facts as extraction units improves the performance on ROUGE-2 and ROUGE-L, since it provides more accurate alignment on the salient information and reduces the proportion of meaningless words in summaries. After imposing the structure information on BERT, we further develop the potential of fact-level extractive summarization. All three variants of our model achieve improvement on ROUGE scores. Among the variants, the model with additional sentence-level information performs best. This is not surprising at all. In fact, sentence-level relationships contain the rich contextual information. Unexpectedly, we find that it does not lead to the best result by combining all three levels of information. We suspect that it may be due to that the document-level information is not effective enough for single document summarization.

\begin{table}
	\centering
	\begin{tabular}{lccc}
		\hline \textbf{Model} & \textbf{ROUGE-1} & \textbf{ROUGE-2} & \textbf{ROUGE-L} \\ \hline
		Lead-3 (our) & 40.25 & 17.40 & 36.41 \\
		Lead-3 \cite{liu2019text} & 40.43 & 17.62 &36.67 \\
		\hline
		SummaRunner \cite{nallapati2017summarunner} & 39.60 & 16.20 & 35.30 \\
		BANDITSUM \cite{dong2018banditsum} &41.50 &18.70 &37.60 \\
		NEUSUM \cite{zhou2018neural} & 41.59 & 19.01 & 37.98 \\
		HIBERT \cite{zhang2019hibert} & 42.37 & 19.95 & 38.83 \\
		PNBERT+RL \cite{zhong2019searching} & 42.69 & 19.60 & 38.85 \\
		BERTEXT+RL \cite{bae2019summary} & 42.76 & 19.87 & 39.11 \\
		BERTSUM \cite{liu2019text} & 43.23 & 20.22 & 39.60 \\
		\hline
		BERTSUM (our) & 42.78 & 19.79 & 39.03 \\
		BERTSUM+fact & 42.81 & 19.93 & 39.84 \\     
		Our d+f & 42.97 & 20.09 & 39.98 \\   
		Our s+f & 43.10 & 20.17 & 40.10 \\   
		Our d+s+f & 43.01 & 20.07 & 40.02 \\		
		\hline
	\end{tabular}
	\caption{\label{font-table}Results on CNN/DM test set. BERTSUM+fact refers to the BERTSUM baseline which simply changes the primary textual unit for extraction from sentences to facts. The last three models represent three variants of our method. For example, d+f means that we use document-level information and fact-level information for summarization}
\end{table}

\subsection{Human Evaluation}
As we know, although the ROUGE metric has long been regarded as a classical evaluation metric in summarization, it does not always reflect the true quality of summaries. Hence, we also conduct human evaluation to check the fact-level recall, precision and F1-score from 50 random samples. As shown in Table 4, compared with sentence extraction, precision of fact extraction increases from 32\% to 39\%, which demonstrates the ability of our model to reduce the over-extraction problem. As for recall, we also gain some degree of improvement. 

\begin{table}[H]
	\centering
	\begin{tabular}{lccc}
		\hline \textbf{Model} & \textbf{Recall} & \textbf{Precision} & \textbf{F1} \\ \hline
		BERTSUM+fact & 41.79 & 38.78 & 39.52 \\
		BERTSUM & 40.45 & 31.79 & 34.98 \\
		\hline
	\end{tabular}
	\caption{\label{font-table}Human evaluation results on CNN/DM test set.}
\end{table}

\subsection{Model Analysis}
\paragraph{Position of Extracted Facts:}
In addition to evaluating the models through ROUGE, we further look into the details of our model. As we know, in a document the sentences at the beginning tend to be more important and are much more likely to be included in summaries. That is the reason why lead-3 is such a strong baseline in the CNN/DailyMail dataset. It turns out that current models may rely too much on the positional information. We feel it is important to analyze the position of the selected units in the source text. To ensure fairness in comparison, we convert all sentences into their corresponding fact units in this experiment. Table 5 shows the proportion of the selected facts that appear in the source text at different fact positions. 1-5 refers to the first five positions, 6-10 represents the positions 6 to 10, and “Rest” means the rest of the positions except 1-15. The results are obtained from the oracle summaries generated by sentence extraction, the oracle summaries generated by facts, those generated by BERTSUM (truncated to four facts) and our s+f. We find that the oracle summaries are distributed smoothly across documents , while the summaries generated by both models highly bias towards the beginning text. The difference is that our model shows a flatter curve on the first 10 facts, indicating that our model can achieve better diversity. 

\begin{table}[H]
	\centering
	\begin{tabular}{lcccc}
		\hline \textbf{Model} & \textbf{1-5} & \textbf{6-10} & \textbf{11-15} & \textbf{Rest} \\ \hline
		Oracle s & 33.11 & 23.15 & 11.79 & 31.95\\
		Oracle f & 29.34 & 21.81 & 11.89 & 36.96\\
		BERTSUM & 59.39 & 26.65 & 6.12 & 7.84\\
		Our s+f & 55.66 & 30.33 & 8.15 & 5.86\\
		
		\hline
	\end{tabular}
	\caption{\label{font-table3}Proportion of extracted facts according to their position in the original text.}
\end{table}

\paragraph{Ablation Study:}
Since the transformer-based model provides us an effective way to disentangle the information from different sources, we are allowed to design an experiment to investigate what roles these types of information play in our model. As shown in Table 6, with our s+f as a base model, we report the ROUGE score after removing segment embeddings or positional embeddings from it. We can observe that the performance of the model without segment embeddings decreases to the same level as BERTSUM+Fact, since it is challenging for a model to understand the hierarchical structure when there is no difference among the information from different sources. We believe that our segmentation strategy is essential for learning the structural information. As for positional embeddings, it is not a surprise that the performance drops by a large margin. According to \cite{zhong2019searching}, current extractive models heavily rely on positional information. The bottom half of Table 6 contains the performance of other models without positional information, including Transformer from \cite{zhong2019searching} and BERTSUM from our experiments. Compared to these models, the hierarchical structure leads to notable improvement, demonstrating that our model has good potential to learn more semantic information rather than simply relying on the positional information. 

\begin{table}[H]
	\centering
	\begin{tabular}{lccc}
		\hline \textbf{Model} & \textbf{R-1} & \textbf{R-2} & \textbf{R-L} \\ \hline
		Original & 43.10  & 20.17 & 40.10\\ 
		- Segment & 42.91  & 19.98 & 39.87\\
		- Position & 39.55 & 17.37 & 36.52 \\ \hline
		Transformer & 37.90 & 15.69 & 34.31 \\
		BERTSUM & 37.97 & 15.93 & 34.66 \\
		\hline
	\end{tabular}
	\caption{\label{font-table4}Results for disentangling test on CNN/DM test set.}
\end{table}

\subsection{Case Study}

Table 7 illustrates an example to show why facts are more effective than sentences in extractive summarization. Here, we use a semicolon to separate different facts in one sentence. The words in italics refer to the sentences selected by the greedy algorithm used in previous models and the colored text represents the facts selected by fact-level alignment, where each selected fact corresponds to one fact in a gold summary. Based on our observation, one sentence may contain several facts but only one of them is included in summary. Meanwhile, although the first sentence selected by the greedy algorithm has a high ROUGE score against the gold summary, it does not correlate with any important facts. On the contrary, our model proves its ability to align the gold summary to its correct location in the source text. The last part of the table contains a summary generated by our model. We can see that our model successfully extracts two gold facts out of four.As a contrast, we also mark the sentences generated by BERTSUM using underscore. It is observed that BERTSUM tends to focus on several leading sentences but only one of them actually contains a gold fact.

\begin{table}[t]
	\centering
	\begin{tabular}{p{15cm}}
		\hline \textbf{Document} \\ \hline
		\underline{Virtual reality may still seem like a hobby reserved for hardcore gamers; but as headsets drop in}
		\underline{price it is on the verge of becoming mainstream.}
		One firm helping to fuel this trend is immerse.
		\underline{\textit{It has created a virtual reality headset that works with any android and ios phone; is compatible}} \underline{\textit{with hundreds of virtual reality apps from the respective stores.}}
		\underline{\textcolor[rgb]{1,0,0}{The immerse virtual reality headset}}
		\underline{\textcolor[rgb]{1,0,0}{is available from firebox for 29.}}
		\textcolor[rgb]{0.3,0.7,0.3}{It works with any android and ios phone that can run virtual reality apps from the respective stores and play any 3d movie.}
		\textcolor[rgb]{0,0,1}{The maximum size of compatible devices is 3;}which means it will work with the iphone 6 ; not the iphone 6 plus , for example. ... 
		\textit{\textcolor[rgb]{0.5,0,1}{Immerse calls itself an affordable alternative to rivals such as oculus rift;} which is expected to launch a consumer version soon with prices ranging from between 200 and 400.}
		\textit{Immerse is available to buy from firebox and can be shipped internationally.}\\
		
		\hline \textbf{Gold summary} \\ \hline
		\textcolor[rgb]{1,0,0}{1: The immerse virtual reality headset is available from firebox for 29.}\\
		\textcolor[rgb]{0.3,0.7,0.3}{2: It works with android and ios phones via virtual reality apps and 3d films.}\\
		\textcolor[rgb]{0,0,1}{3: The maximum size of the device must be 3.}\\
		\textcolor[rgb]{0.5,0,1}{4: It calls itself an affordable alternative to rivals such as oculus rift.}\\	\hline
		
		\hline \textbf{Our s+f} \\ \hline		
		The immerse virtual reality headset is available from firebox for 29.	\\	
		The maximum size of the device must be 3.\\
		Is compatible with hundreds of virtual reality apps from the respective stores.	\\
		Virtual reality may still seem like a hobby reserved for hardcore gamers.\\	\hline
		
	\end{tabular}
	\caption{\label{font-table5}Example from the CNN/DailMail test dataset}
\end{table}

\section{Conclusion}
In this paper, we propose to extract fact-level semantic units for extractive summarization. Without increasing the scale of the model, we propose a hierarchical graph mask on BERT to fully utilize the structural information among different semantic levels. Experiments on the CNN/DaliyMail dataset shows that our model achieves state-of-the-art results. In the future, we would like to use our current model to guide abstractive summarization.

\section*{Acknowledgements}

The work described in this paper was supported by and Research Grants Council of Hong Kong (15203617 and 5210919) and National Natural Science Foundation of China (61672445).

\bibliographystyle{coling}
\bibliography{coling2020}

\begin{thebibliography}{}

\bibitem[\protect\citename{Bae \bgroup et al.\egroup }2019]{bae2019summary}
Sanghwan Bae, Taeuk Kim, Jihoon Kim, and Sang-goo Lee.
\newblock 2019.
\newblock Summary level training of sentence rewriting for abstractive
  summarization.
\newblock {\em arXiv preprint arXiv:1909.08752}.

\bibitem[\protect\citename{Bui \bgroup et al.\egroup }2016]{bui2016extractive}
Duy Duc~An Bui, Guilherme Del~Fiol, John~F Hurdle, and Siddhartha Jonnalagadda.
\newblock 2016.
\newblock Extractive text summarization system to aid data extraction from full
  text in systematic review development.
\newblock {\em Journal of biomedical informatics}, 64:265--272.

\bibitem[\protect\citename{Cao \bgroup et al.\egroup }2018]{cao2018faithful}
Ziqiang Cao, Furu Wei, Wenjie Li, and Sujian Li.
\newblock 2018.
\newblock Faithful to the original: Fact aware neural abstractive
  summarization.
\newblock In {\em Thirty-Second AAAI Conference on Artificial Intelligence}.

\bibitem[\protect\citename{Cheng and Lapata}2016]{cheng2016neural}
Jianpeng Cheng and Mirella Lapata.
\newblock 2016.
\newblock Neural summarization by extracting sentences and words.
\newblock {\em arXiv preprint arXiv:1603.07252}.

\bibitem[\protect\citename{Devlin \bgroup et al.\egroup }2018]{devlin2018bert}
Jacob Devlin, Ming-Wei Chang, Kenton Lee, and Kristina Toutanova.
\newblock 2018.
\newblock Bert: Pre-training of deep bidirectional transformers for language
  understanding.
\newblock {\em arXiv preprint arXiv:1810.04805}.

\bibitem[\protect\citename{Dong \bgroup et al.\egroup }2018]{dong2018banditsum}
Yue Dong, Yikang Shen, Eric Crawford, Herke van Hoof, and Jackie Chi~Kit
  Cheung.
\newblock 2018.
\newblock Banditsum: Extractive summarization as a contextual bandit.
\newblock {\em arXiv preprint arXiv:1809.09672}.

\bibitem[\protect\citename{Dong \bgroup et al.\egroup }2019]{dong2019unified}
Li~Dong, Nan Yang, Wenhui Wang, Furu Wei, Xiaodong Liu, Yu~Wang, Jianfeng Gao,
  Ming Zhou, and Hsiao-Wuen Hon.
\newblock 2019.
\newblock Unified language model pre-training for natural language
  understanding and generation.
\newblock In {\em Advances in Neural Information Processing Systems}, pages
  13042--13054.

\bibitem[\protect\citename{Erkan and Radev}2004]{erkan2004lexrank}
G{\"u}nes Erkan and Dragomir~R Radev.
\newblock 2004.
\newblock Lexrank: Graph-based lexical centrality as salience in text
  summarization.
\newblock {\em Journal of artificial intelligence research}, 22:457--479.

\bibitem[\protect\citename{Fang \bgroup et al.\egroup
  }2019]{fang2019hierarchical}
Yuwei Fang, Siqi Sun, Zhe Gan, Rohit Pillai, Shuohang Wang, and Jingjing Liu.
\newblock 2019.
\newblock Hierarchical graph network for multi-hop question answering.
\newblock {\em arXiv preprint arXiv:1911.03631}.

\bibitem[\protect\citename{Gehrmann \bgroup et al.\egroup
  }2018]{gehrmann2018bottom}
Sebastian Gehrmann, Yuntian Deng, and Alexander~M Rush.
\newblock 2018.
\newblock Bottom-up abstractive summarization.
\newblock {\em arXiv preprint arXiv:1808.10792}.

\bibitem[\protect\citename{Hermann \bgroup et al.\egroup
  }2015]{hermann2015teaching}
Karl~Moritz Hermann, Tomas Kocisky, Edward Grefenstette, Lasse Espeholt, Will
  Kay, Mustafa Suleyman, and Phil Blunsom.
\newblock 2015.
\newblock Teaching machines to read and comprehend.
\newblock In {\em Advances in neural information processing systems}, pages
  1693--1701.

\bibitem[\protect\citename{Koncel-Kedziorski \bgroup et al.\egroup
  }2019]{koncel2019text}
Rik Koncel-Kedziorski, Dhanush Bekal, Yi~Luan, Mirella Lapata, and Hannaneh
  Hajishirzi.
\newblock 2019.
\newblock Text generation from knowledge graphs with graph transformers.
\newblock {\em arXiv preprint arXiv:1904.02342}.

\bibitem[\protect\citename{Lin}2004]{lin-2004-rouge}
Chin-Yew Lin.
\newblock 2004.
\newblock {ROUGE}: A package for automatic evaluation of summaries.
\newblock In {\em Text Summarization Branches Out}, pages 74--81, Barcelona,
  Spain, July. Association for Computational Linguistics.

\bibitem[\protect\citename{Liu and Lapata}2019]{liu2019text}
Yang Liu and Mirella Lapata.
\newblock 2019.
\newblock Text summarization with pretrained encoders.
\newblock {\em arXiv preprint arXiv:1908.08345}.

\bibitem[\protect\citename{Liu \bgroup et al.\egroup }2018]{liu2018toward}
Fei Liu, Jeffrey Flanigan, Sam Thomson, Norman Sadeh, and Noah~A Smith.
\newblock 2018.
\newblock Toward abstractive summarization using semantic representations.
\newblock {\em arXiv preprint arXiv:1805.10399}.

\bibitem[\protect\citename{Liu \bgroup et al.\egroup }2019]{liu2019single}
Yang Liu, Ivan Titov, and Mirella Lapata.
\newblock 2019.
\newblock Single document summarization as tree induction.
\newblock In {\em Proceedings of the 2019 Conference of the North American
  Chapter of the Association for Computational Linguistics: Human Language
  Technologies, Volume 1 (Long and Short Papers)}, pages 1745--1755.

\bibitem[\protect\citename{Mihalcea and Tarau}2004]{mihalcea2004textrank}
Rada Mihalcea and Paul Tarau.
\newblock 2004.
\newblock Textrank: Bringing order into text.
\newblock In {\em Proceedings of the 2004 conference on empirical methods in
  natural language processing}, pages 404--411.

\bibitem[\protect\citename{Nallapati \bgroup et al.\egroup
  }2017]{nallapati2017summarunner}
Ramesh Nallapati, Feifei Zhai, and Bowen Zhou.
\newblock 2017.
\newblock Summarunner: A recurrent neural network based sequence model for
  extractive summarization of documents.
\newblock In {\em Thirty-First AAAI Conference on Artificial Intelligence}.

\bibitem[\protect\citename{Narayan \bgroup et al.\egroup
  }2018]{narayan2018ranking}
Shashi Narayan, Shay~B Cohen, and Mirella Lapata.
\newblock 2018.
\newblock Ranking sentences for extractive summarization with reinforcement
  learning.
\newblock {\em arXiv preprint arXiv:1802.08636}.

\bibitem[\protect\citename{Paulus \bgroup et al.\egroup }2017]{paulus2017deep}
Romain Paulus, Caiming Xiong, and Richard Socher.
\newblock 2017.
\newblock A deep reinforced model for abstractive summarization.
\newblock {\em arXiv preprint arXiv:1705.04304}.

\bibitem[\protect\citename{Pennington \bgroup et al.\egroup
  }2014]{pennington2014glove}
Jeffrey Pennington, Richard Socher, and Christopher~D Manning.
\newblock 2014.
\newblock Glove: Global vectors for word representation.
\newblock In {\em Proceedings of the 2014 conference on empirical methods in
  natural language processing (EMNLP)}, pages 1532--1543.

\bibitem[\protect\citename{Radford \bgroup et al.\egroup
  }2018]{radford2018improving}
Alec Radford, Karthik Narasimhan, Tim Salimans, and Ilya Sutskever.
\newblock 2018.
\newblock Improving language understanding by generative pre-training.
\newblock {\em URL https://s3-us-west-2. amazonaws.
  com/openai-assets/researchcovers/languageunsupervised/language understanding
  paper. pdf}.

\bibitem[\protect\citename{Vaswani \bgroup et al.\egroup
  }2017]{vaswani2017attention}
Ashish Vaswani, Noam Shazeer, Niki Parmar, Jakob Uszkoreit, Llion Jones,
  Aidan~N Gomez, {\L}ukasz Kaiser, and Illia Polosukhin.
\newblock 2017.
\newblock Attention is all you need.
\newblock In {\em Advances in neural information processing systems}, pages
  5998--6008.

\bibitem[\protect\citename{Williams \bgroup et al.\egroup
  }2018]{williams2018latent}
Adina Williams, Andrew Drozdov*, and Samuel~R Bowman.
\newblock 2018.
\newblock Do latent tree learning models identify meaningful structure in
  sentences?
\newblock {\em Transactions of the Association for Computational Linguistics},
  6:253--267.

\bibitem[\protect\citename{Zhang \bgroup et al.\egroup }2018]{zhang2018neural}
Xingxing Zhang, Mirella Lapata, Furu Wei, and Ming Zhou.
\newblock 2018.
\newblock Neural latent extractive document summarization.
\newblock {\em arXiv preprint arXiv:1808.07187}.

\bibitem[\protect\citename{Zhang \bgroup et al.\egroup }2019]{zhang2019hibert}
Xingxing Zhang, Furu Wei, and Ming Zhou.
\newblock 2019.
\newblock Hibert: Document level pre-training of hierarchical bidirectional
  transformers for document summarization.
\newblock {\em arXiv preprint arXiv:1905.06566}.

\bibitem[\protect\citename{Zhong \bgroup et al.\egroup
  }2019]{zhong2019searching}
Ming Zhong, Pengfei Liu, Danqing Wang, Xipeng Qiu, and Xuanjing Huang.
\newblock 2019.
\newblock Searching for effective neural extractive summarization: What works
  and what's next.
\newblock {\em arXiv preprint arXiv:1907.03491}.

\bibitem[\protect\citename{Zhou \bgroup et al.\egroup }2018]{zhou2018neural}
Qingyu Zhou, Nan Yang, Furu Wei, Shaohan Huang, Ming Zhou, and Tiejun Zhao.
\newblock 2018.
\newblock Neural document summarization by jointly learning to score and select
  sentences.
\newblock {\em arXiv preprint arXiv:1807.02305}.

\end{thebibliography}

\end{document}